\documentclass[conference]{IEEEtran}
\IEEEoverridecommandlockouts 

\usepackage[utf8]{inputenc}
\usepackage[T1]{fontenc}
\usepackage{url}
\usepackage{booktabs}
\usepackage{amsfonts}
\usepackage{amsmath}
\usepackage{graphicx}
\usepackage{xcolor}
\usepackage{multirow}
\usepackage{cite}
\usepackage{balance}
\usepackage{hyperref}
\hypersetup{hidelinks}

\begin{document}

\title{
How Small Can You Go? LoRA Fine-Tuning\\270M--8B Models for Merchant Information Extraction in Financial Transactions
}
%

\author{
\IEEEauthorblockN{Donghao Huang\textsuperscript{1,3}, Tom\'a\v{s} Drietomsk\'y\textsuperscript{2}, Benjamin Barrett\textsuperscript{1}, Zhaoxia Wang\textsuperscript{3,*}}
\thanks{*Corresponding author: zxwang@smu.edu.sg}
\IEEEauthorblockA{\textsuperscript{1}Research and Development, Mastercard, Arlington, VA, USA\\
\textsuperscript{2}Economic Intelligence, Mastercard, Prague, Czech Republic\\
\textsuperscript{3}School of Computing and Information Systems, Singapore Management University, Singapore\\
\{donghao.huang, tomas.drietomsky, benjamin.barrett\}@mastercard.com; zxwang@smu.edu.sg}
}

\maketitle
\newcommand{\icdmpubid}{}
\newcommand{\icdmpubidunset}{}
\ifx\icdmpubid\icdmpubidunset
  \newcommand{\icdmpubidadjcol}{}%
  \typeout{^^JWARNING: IEEE copyright notice not set. Define \string\icdmpubid\space before camera-ready submission.^^J}%
\else
  \newcommand{\icdmpubidadjcol}{\IEEEpubidadjcol}%
  \IEEEpubid{\makebox[\columnwidth]{\icdmpubid\hfill}%
    \hspace{\columnsep}\makebox[\columnwidth]{}}%
\fi

\begin{abstract}
Merchant information extraction turns noisy financial transaction descriptors into structured fields at production scale. Our deployed LoRA-fine-tuned LLaMA~3.1-8B reaches 96.95\% F1, but its memory and throughput motivate smaller replacements. We evaluate 23 retained fine-tuning runs plus a separately trained production reference, spanning Gemma~3 (270M--4B), Qwen~3.5 (0.8B--4B), Aya~3.35B, and LLaMA~3.1-8B across LoRA ranks, prompts, training templates, and serving environments. A rank-8 LLaMA fine-tune reaches 96.75\% F1, only 0.20 points below the rank-32 production reference. Qwen~3.5~4B with JSON-Only prompting reaches 96.60\% F1 and strict record-level exact match of 91.67\%, with a $1.2\times$ lower inverse-throughput time estimate than the rank-8 8B model. Qwen~3.5~0.8B reaches 94.75\% F1, and Qwen Think and Nothink templates stay within 0.004 F1. Matched no-adapter baselines for all 16 model--prompt pairs trail their fine-tuned counterparts by 20.8--92.1 F1 points. Across 14 Databricks endpoints, mean F1 change from local evaluation is $-0.0081$; Aya is the only family with a 2.7--5.1 point decline. These results show that compact fine-tuned models can preserve most extraction accuracy, but model selection must account for prompt choice, throughput, and serving-stack behavior.
\end{abstract}


\section{Introduction}

\icdmpubidadjcol
Financial transaction processing generates billions of text strings daily, each encoding merchant identity, location, and transaction metadata in compressed, noisy formats. Transaction descriptors may also contain dynamic, transaction-specific components or intermediary identifiers. A representative transaction string such as \texttt{``VIATOR IT-1558003355 360 3RD ST, STE 400, SAN FRANCISCO 7027495744''} may populate several of the ten entity fields---merchant name, intermediary identifiers, dynamic components, location, and contact fields (Table~\ref{tab:entity_fields}). Accurate extraction is critical for fraud detection, merchant categorization, and consumer analytics~\cite{du2025nlpfinance}.

This study positions merchant information extraction as an upstream analytical component for Mastercard products and algorithms that operate on Mastercard-derived transaction descriptor data. The proposed approach is not designed to authorize, route, clear, settle, or otherwise process Mastercard payment transactions; rather, it operates on textual descriptors from existing transactions to improve merchant understanding, data enrichment and analytics, downstream product quality, and the overall reliability of merchant intelligence systems.

Our current production system uses a LoRA-fine-tuned LLaMA~3.1-8B~\cite{dubey2024llama} that achieves approximately 97\% F1 on this task. However, deploying an 8B-parameter model raises practical concerns: high GPU memory requirements (16+ GB in half-precision), limited inference throughput, and substantial energy costs~\cite{strubell2019energy}. As the system must handle high transaction traffic, even small improvements in efficiency can translate into meaningful operational savings.

The proliferation of capable small language models (SLMs)---Gemma~3~\cite{team2024gemma}, Qwen~3.5~\cite{yang2024qwen2}, Aya~\cite{ustun2024aya}---motivates a systematic investigation: \emph{what is the minimum model size that maintains production-grade accuracy for structured entity extraction?}

This paper addresses this question through a deployment-oriented empirical study. Rather than proposing a new adaptation algorithm, we provide a systematic comparison spanning 23 retained fine-tuning runs plus a separately trained production reference across four model families, two prompting strategies, and two training templates. Our contributions are:

\begin{enumerate}
    \item A \textbf{matched zero-shot baseline sweep} over all eight base models under both prompts, bounding the adaptation effect at 20.8--92.1 F1 points and the best no-adapter configuration at 0.758 F1 (Section~\ref{sec:zeroshot}).
    \item A \textbf{rank and scale study} from 0.27B to 8B showing that rank~8 LLaMA~3.1-8B is within 0.20 F1 points of the rank-32 production reference and Qwen~3.5~4B is within 0.35 points at half the model size (Section~\ref{sec:scaling}).
    \item A \textbf{prompt and template comparison} showing when Free-Thinking or JSON-Only instructions help and that Qwen Think and Nothink training remains within 0.004 F1 (Section~\ref{sec:prompt}).
    \item An \textbf{accuracy--throughput analysis} showing that Qwen~3.5~4B~JO trades 0.15 F1 points for a $1.2\times$ lower effective seconds-per-sample estimate, computed from median batched throughput, relative to rank-8 LLaMA~3.1-8B~FT (Section~\ref{sec:pareto}).
    \item A \textbf{managed-serving validation} across 14 sub-8B endpoints, identifying Aya as the sole family with a material serving-time accuracy gap (Section~\ref{sec:serving}).
    \item \textbf{Concrete deployment guidance} by accuracy tier, prompt, template, adapter rank, and serving compatibility (Section~\ref{sec:deployment}).
\end{enumerate}

The remainder of the paper is organized as follows. Section~\ref{sec:related} reviews related work; Section~\ref{sec:methodology} defines the task, data, models, training, and metrics; Section~\ref{sec:results} presents local evaluation; Section~\ref{sec:serving} validates managed endpoints; Section~\ref{sec:deployment} gives deployment recommendations; Section~\ref{sec:limitations} discusses limitations and generalizability; and Section~\ref{sec:conclusion} concludes.

\section{Related Work}
\label{sec:related}

\subsection{Parameter-Efficient Fine-Tuning}

LoRA~\cite{hu2022lora} introduced low-rank adaptation as a parameter-efficient alternative to full fine-tuning, and has since been extended by QLoRA~\cite{dettmers2023qlora} for quantized training and AdaLoRA~\cite{zhang2023adalora} for adaptive rank. In practice these methods are most readily applied through open-source toolkits, and LLaMA-Factory~\cite{zheng2024llamafactory} (ACL~2024) is among the most widely adopted: it provides a unified pipeline for parameter-efficient fine-tuning of 100+ open-source LLMs and VLMs, supports a range of training objectives (SFT, DPO, KTO, ORPO, PPO), and typically adds support for new model releases on a day-zero basis. Its widespread adoption has produced a growing body of follow-on research---for example, LoRA fine-tuning of 3.5B--72B open-source LLMs for Chinese-to-English translation~\cite{huang2025chinese} and few-shot prompting combined with LoRA fine-tuning of LLMs for logical reasoning on Turtle Soup puzzles~\cite{huang2025turtle}---and we adopt it as the fine-tuning backbone of this study.

\subsection{Small Language Models}

Models such as Phi-3~\cite{abdin2024phi3}, Gemma~3~\cite{team2024gemma}, and Qwen~3.5~\cite{yang2024qwen2} have demonstrated strong performance at 1--4B parameters. Qwen~3.5 introduces a hybrid architecture combining Mamba state-space layers~\cite{gu2023mamba} with transformer attention~\cite{dao2024transformers}. Concrete production-oriented evaluations are starting to show that small open-source LMs can match much larger proprietary models on demanding tasks: in LLM-based agentic payment systems, for example, Gemma~4~E4B has been shown to match GPT-5.2's perfect Agentic Success Rate (ASR) despite being orders of magnitude smaller~\cite{huang2026payments}.

\subsection{
Merchant Information Extraction in Financial Transactions
}
%
Traditional Named Entity Recognition (NER) approaches rely on CRF-based sequence labeling methods~\cite{lample2016neural, ma2016end}. Recent work has reframed information extraction as a generation task~\cite{wang2023gptner, li2023evaluating}. Financial NER tasks, such as merchant information extraction from noisy bank transaction strings, present unique challenges due to abbreviated text, limited context, and non-standard formatting~\cite{du2025nlpfinance, shah2022flue}. Chain-of-thought prompting~\cite{wei2022chain} has also been shown to improve performance on structured prediction tasks~\cite{zhang2023multimodal}.

CRF~\cite{lafferty2001conditional, lample2016neural} and encoder-based~\cite{devlin2019bert} sequence labelers are important task-specific alternatives, but the present study asks a narrower deployment question: how small a model can replace an existing generative LLaMA system while preserving its ten-key JSON contract. Our supervision consists of field-value objects rather than token-level BIO labels, and rules such as removing duplicate substrings from \textsc{Merchant\_Name} and reassigning them to another field require non-trivial alignment and post-processing in a sequence-labeling formulation.

\section{Methodology}
\label{sec:methodology}

\subsection{Task Definition}

Given a raw transaction descriptor $x$, the system produces a JSON object $y$ with ten entity fields (Table~\ref{tab:entity_fields}). Target values are derived from $x$, sometimes with expert normalization, and are not required to be literal substrings; absent fields are empty.

\begin{table}[t]
\centering
\caption{Entity fields and dataset characteristics (8,015 samples).}
\label{tab:entity_fields}
\footnotesize
\begin{tabular}{@{}lrrl@{}}
\toprule
\textbf{Field} & \textbf{Present (\%)} & \textbf{Unique} & \textbf{Diff.} \\
\midrule
\textsc{Merchant\_Name} & 100.0 & 6,411 & Hard \\
\textsc{State} & 68.5 & 57 & Easy \\
\textsc{City} & 46.8 & 1,799 & Med. \\
\textsc{Street} & 31.2 & 2,070 & Med. \\
\textsc{Dyn.\_Descriptor} & 29.1 & 2,278 & Hard \\
\textsc{Intermediary} & 26.7 & 457 & Med. \\
\textsc{Phone} & 19.6 & 947 & Easy \\
\textsc{Country} & 12.7 & 4 & Hard$^\dagger$ \\
\textsc{Zip} & 12.0 & 843 & Easy \\
\textsc{Website} & 5.4 & 170 & Easy \\
\bottomrule
\multicolumn{4}{l}{\footnotesize $^\dagger$Low frequency despite few classes.}
\end{tabular}
\end{table}

\subsection{Dataset}
\label{sec:dataset}

The dataset contains 8,015 transaction descriptors derived from Mastercard transaction data. Human experts created the ground-truth values for all ten fields. The data are split into 6,125 training, 668 validation, and 1,222 test samples. All reported results use the 1,213 test records for which the production reference retained predictions. Descriptor length is 7--86 characters (median 34, mean 35.5) and 1--16 whitespace-delimited tokens (median 5, mean 5.4). As shown in Table~\ref{tab:entity_fields}, entity fields exhibit highly imbalanced presence rates (100\% for \textsc{Merchant\_Name} to 5.4\% for \textsc{Website}).


\subsection{Models}

We evaluate four model families, all LoRA fine-tuned in this study:

\begin{itemize}
    \item \textbf{Gemma~3}~\cite{team2024gemma}: 270M, 1B, 4B (instruction-tuned, standard transformer).
    \item \textbf{Qwen~3.5}~\cite{yang2024qwen2}: 0.8B, 2B, 4B (hybrid Mamba-attention architecture with native extended thinking support).
    \item \textbf{Aya}~\cite{ustun2024aya}: 3.35B (\texttt{tiny-aya-global}, Cohere2 architecture).
    \item \textbf{LLaMA~3.1-8B}~\cite{dubey2024llama}: 8B (instruction-tuned, standard transformer). Re-fine-tuned with LoRA rank~8 to enable direct comparison against the production baseline.
\end{itemize}

\noindent Our production baseline is a separately trained LLaMA~3.1-8B attributed in internal records to LoRA rank~32 (F1=96.95\%), included as a fixed reference point. The retained parsed-output artifact establishes its predictions and aggregate metrics but does not retain the adapter, checkpoint, model revision, or inference configuration; rank~32 is therefore author-attested rather than independently reconstructible. We additionally fine-tune LLaMA~3.1-8B with rank~8 under our standardized pipeline so that the 8B model can be compared head-to-head against the smaller candidates in this study.

\subsection{Fine-Tuning Pipeline}

All eight models (including our LLaMA~3.1-8B rank-8 fine-tunes) are trained with LLaMA-Factory~\cite{zheng2024llamafactory} on an NVIDIA DGX Spark---a compact desktop workstation powered by the GB10 Blackwell Superchip with 128\,GB unified LPDDR5x memory. Both fine-tuning and all reported throughput measurements use this single device. The zero-shot baselines of Section~\ref{sec:zeroshot} were run on a separate 16\,GB RTX~4090 under the same software stack, except the two LLaMA~3.1-8B rows, which exceed that card in bfloat16 and were run on the DGX Spark.

This choice deliberately separates experimentation from production. Governed production environments apply controlled model-onboarding policies, and managed fine-tuning stacks prioritize stability over day-zero coverage of newly released architectures. LLaMA-Factory tracks new releases closely, so running it on a self-contained DGX Spark outside the production perimeter lets us train, evaluate, and select among candidate base models as soon as their weights are available. It is also materially cheaper: the 23 retained training runs totaled $\approx$462~GPU-hours of training plus 195~hours of checkpoint evaluation (12~checkpoints per run; 265 retained checkpoint evaluations)---$\approx$657~GPU-hours in all. Only the best-of-family adapters identified through this sweep are subsequently promoted as managed serving endpoints---a two-stage \emph{experiment-then-deploy} workflow whose viability we confirm via the endpoint validation in Section~\ref{sec:serving}.

Key training hyperparameters (full configuration in Appendix~\ref{app:training}):

\begin{itemize}
    \item \textbf{LoRA}: rank $r{=}8$, $\alpha{=}16$ (scaling $\alpha/r{=}2.0$)
    \item \textbf{Training}: 6 epochs (2,298 steps), LR $10^{-4}$, cosine schedule; Aya rank~32 alone uses $5\times10^{-5}$
    \item \textbf{Batch}: Effective size 16 (4 $\times$ 4 gradient accumulation)
    \item \textbf{Checkpoints}: Every 200 steps (12 per model)
\end{itemize}

\subsection{Prompt Strategies}

We compare two compound prompt variants with identical JSON output labels (full templates in Appendix~\ref{app:prompts}):

\textbf{Free-Thinking (FT):} A two-step instruction---first analyze internally, then apply entity rules and output JSON---plus an explicit ten-key output schema. Prompt: 3,700 chars.

\textbf{JSON-Only (JO):} Direct extraction with the same rules, no internal-analysis instruction, and no literal JSON schema. Prompt: 2,870 chars.

\subsection{Evaluation Protocol}

For each of the 22 retained runs with a complete 12-checkpoint sweep, we report the checkpoint with the highest F1 on the common 1,213-record evaluation subset (Section~\ref{sec:dataset}). This test-selected checkpoint is an oracle characterization of each training trajectory, not a deployable selection procedure, and can optimistically bias the reported maxima. Aya rank~32 contributes only its retained checkpoint, and the separately trained production reference is fixed rather than checkpoint-selected in this study. For field-level scoring, predicted and reference values are canonicalized by uppercasing and removing spaces, asterisks, commas, and hyphens; a field matches only when the two canonicalized strings are identical. Precision, recall, and F1 use these canonicalized exact field matches, and overall F1 is micro-averaged across all ten fields. Micro-F1 is preferable to raw field accuracy because the fields are optional and highly imbalanced: it distinguishes spurious non-empty predictions and missed values rather than rewarding the many true empty fields. We additionally report strict record-level exact match (R-EM), the fraction of records whose ten raw values match exactly. Missing keys are scored as empty strings for F1 and R-EM; schema-valid outputs instead contain exactly the ten required string-valued keys.
All local evaluation runs in bfloat16 without quantization. Inference throughput is measured on an NVIDIA DGX Spark with batch size~4 and deterministic decoding ($\texttt{do\_sample}{=}\texttt{false}$, $T{=}0$). Every reported speed is the median samples per second over three interleaved full-test-set evaluations of that configuration's selected checkpoint, under one protocol and one software stack, with predictions bit-identical across repetitions in every case. The Pareto plot uses the inverse of that median as an effective seconds-per-sample estimate, which is not single-request latency.

\section{Experimental Results}
\label{sec:results}

\subsection{Overall Performance}
\label{sec:overall}

Table~\ref{tab:main_results} presents overall performance. Every study row is LoRA-fine-tuned; the separately trained production model is included only as a fixed reference and its rank~32 attribution is author-attested. That reference achieves 96.95\% F1 and 92.42\% strict R-EM (1,121 of 1,213 records), as verified from its retained parsed outputs; the planned public snapshot will include aggregate and match-vector hashes, but not the record-level production outputs. Our rank-8 LLaMA~3.1-8B fine-tune reaches 96.75\% F1 (FT) and 96.57\% (JO)---within 0.20 and 0.38 points of the reference, respectively. The best sub-8B model is Qwen~3.5~4B with JSON-Only prompting at 96.60\% F1 and 91.67\% strict R-EM, a gap of only 0.35 F1 points from the reference at half the parameter count. Gemma~3~4B (FT) and both Qwen~3.5~4B Nothink variants (JO~NT, FT~NT) tie at 96.24\% F1 at closely matched throughput (0.59--0.63 samples/s). R-EM is lower in absolute terms because a record receives credit only when all ten raw fields match, but it supports the same practical conclusion: 4B candidates approach the 8B reference on complete structured outputs. Throughput separates the smaller models sharply but compresses at the top of the range: Gemma~3~270M~FT reaches $6.7\times$ the throughput of Qwen~3.5~4B~JO, which in turn reaches only $1.2\times$ that of LLaMA~3.1-8B~FT. The case for replacing the 8B therefore rests more on memory footprint than on throughput; Section~\ref{sec:pareto} analyzes the trade-off directly.

\begin{table}[t]
\centering
\caption{Canonicalized field metrics and strict R-EM on the 1,213-record common set. Every study row is a complete sweep reported at its oracle-best test checkpoint; the production reference is fixed rather than checkpoint-selected. Speed is the median of repeated batch-4 evaluations of the selected checkpoint on one DGX Spark. NT = Nothink. $^\ast$Production rank-32 reference. $^\ddagger$Our rank-8 fine-tune.}
\label{tab:main_results}
\scriptsize
\setlength{\tabcolsep}{2.7pt}
\begin{tabular}{@{}llrrrrr@{}}
\toprule
\textbf{Model} & \textbf{Pr.} & \textbf{Step} & \textbf{F1} & \textbf{R-EM} & \textbf{P / R} & \textbf{s/s} \\
\midrule
LLaMA 3.1-8B$^\ast$ & FT & --- & .9695 & .9242 & .971/.968 & --- \\
\midrule
LLaMA 3.1-8B$^\ddagger$ & FT & 2000 & .9675 & .9151 & .967/.968 & 0.50 \\
Qwen 3.5 4B & JO & 2000 & \textbf{.9660} & \textbf{.9167} & .967/.965 & 0.61 \\
LLaMA 3.1-8B$^\ddagger$ & JO & 1000 & .9657 & .9134 & .967/.964 & 0.54 \\
Gemma 3 4B & FT & 2000 & .9624 & .9002 & .963/.962 & 0.63 \\
Qwen 3.5 4B & JO NT & 1400 & .9624 & .9035 & .963/.962 & 0.62 \\
Qwen 3.5 4B & FT NT & 1600 & .9624 & .8994 & .963/.962 & 0.59 \\
Qwen 3.5 4B & FT & 1400 & .9615 & .9019 & .962/.961 & 0.55 \\
Aya 3.35B & FT & 1600 & .9572 & .8937 & .956/.958 & 0.74 \\
Gemma 3 4B & JO & 1600 & .9540 & .8829 & .954/.954 & 0.68 \\
Aya 3.35B & JO & 1400 & .9522 & .8813 & .950/.954 & 0.81 \\
Qwen 3.5 2B & FT NT & 1400 & .9518 & .8961 & .951/.953 & 1.27 \\
Qwen 3.5 2B & FT & 1600 & .9510 & .8904 & .948/.954 & 1.21 \\
Qwen 3.5 2B & JO NT & 1600 & .9486 & .8871 & .948/.949 & 1.34 \\
Qwen 3.5 0.8B & FT NT & 1400 & .9475 & .8755 & .948/.947 & 1.86 \\
Qwen 3.5 2B & JO & 1400 & .9464 & .8821 & .948/.945 & 1.28 \\
Qwen 3.5 0.8B & JO & 1200 & .9461 & .8739 & .947/.945 & 1.88 \\
Qwen 3.5 0.8B & FT & 1600 & .9458 & .8730 & .949/.943 & 1.73 \\
Qwen 3.5 0.8B & JO NT & 1600 & .9452 & .8747 & .946/.944 & 2.01 \\
Gemma 3 1B & FT & 1600 & .9393 & .8516 & .942/.937 & 2.23 \\
Gemma 3 1B & JO & 1600 & .9292 & .8293 & .932/.926 & 2.28 \\
Gemma 3 270M & FT & 1800 & .8735 & .7312 & .884/.863 & 4.09 \\
Gemma 3 270M & JO & 2000 & .8553 & .7148 & .870/.842 & 2.78 \\
\bottomrule
\end{tabular}
\end{table}

\subsection{Zero-Shot Baselines}
\label{sec:zeroshot}

To quantify how much of the reported accuracy comes from fine-tuning rather than from the base models, we evaluate each base instruction model without an adapter under both prompts, holding decoding settings, evaluator, and the 1,213-record common subset fixed. Table~\ref{tab:zeroshot} reports the complete 16-run zero-shot sweep and compares each run with the fine-tuned checkpoint using the same prompt.

\begin{table}[t]
\centering
\caption{Zero-shot (no adapter) accuracy and improvement after fine-tuning under the matching prompt, on the 1,213-record common subset. $\Delta$F1 is fine-tuned minus zero-shot F1.}
\label{tab:zeroshot}
\scriptsize
\setlength{\tabcolsep}{2.2pt}
\begin{tabular}{@{}llrrrrrr@{}}
\toprule
\textbf{Model} & \textbf{Pr.} & \textbf{ZS F1} & \textbf{ZS R-EM} & \textbf{Parse} & \textbf{Schema} & \textbf{FT F1} & \textbf{$\Delta$F1} \\
\midrule
Aya 3.35B      & FT & .3839 & .0305 & 0.882 & 0.877 & .9572 & $+$.5733 \\
Aya 3.35B      & JO & .3659 & .0074 & 0.138 & 0.103 & .9522 & $+$.5863 \\
Gemma 3 270M   & FT & .0000 & .0000 & 0.000 & 0.000 & .8735 & $+$.8735 \\
Gemma 3 270M   & JO & .0051 & .0000 & 0.917 & 0.000 & .8553 & $+$.8502 \\
Gemma 3 1B     & FT & .2565 & .0000 & 0.998 & 0.990 & .9393 & $+$.6828 \\
Gemma 3 1B     & JO & .0957 & .0049 & 0.998 & 0.002 & .9292 & $+$.8335 \\
Gemma 3 4B     & FT & .5905 & .1896 & 1.000 & 1.000 & .9624 & $+$.3719 \\
Gemma 3 4B     & JO & .6212 & .2490 & 1.000 & 0.999 & .9540 & $+$.3328 \\
Qwen 3.5 0.8B  & FT & .0248 & .0025 & 1.000 & 1.000 & .9458 & $+$.9210 \\
Qwen 3.5 0.8B  & JO & .2706 & .0775 & 1.000 & 0.000 & .9461 & $+$.6755 \\
Qwen 3.5 2B    & FT & .4866 & .1641 & 1.000 & 0.752 & .9510 & $+$.4644 \\
Qwen 3.5 2B    & JO & .5505 & .2242 & 1.000 & 0.000 & .9464 & $+$.3959 \\
Qwen 3.5 4B    & FT & .6543 & .3693 & 0.740 & 0.681 & .9615 & $+$.3072 \\
Qwen 3.5 4B    & JO & .7583 & .4732 & 1.000 & 0.040 & .9660 & $+$.2077 \\
LLaMA 3.1-8B   & FT & .5798 & .2135 & 1.000 & 0.975 & .9675 & $+$.3877 \\
LLaMA 3.1-8B   & JO & .6980 & .3578 & 0.998 & 0.181 & .9657 & $+$.2677 \\
\bottomrule
\end{tabular}
\end{table}

All 16 matched pairs improve substantially after fine-tuning, from $+$0.208 F1 for Qwen~3.5~4B~JO to $+$0.921 for Qwen~3.5~0.8B~FT. The best zero-shot configuration, Qwen~3.5~4B~JO, reaches 0.758 F1 and 0.473 strict R-EM; its fine-tuned counterpart reaches 0.966 F1 and 0.917 strict R-EM.

Zero-shot accuracy is prompt-sensitive: JO beats FT for six of eight models, while FT leads for Aya~3.35B and Gemma~3~1B. Schema validity requires all ten string-valued keys, whereas micro-F1 scores non-empty fields and treats omitted keys as empty; partial objects can therefore earn F1 despite being schema-invalid. All three Qwen JO runs parse every record, but only 0.0\%, 0.0\%, and 4.0\% are schema-valid; Qwen~3.5~4B nonetheless correctly extracts 705 state, 568 city, and 367 ZIP values. LLaMA~3.1-8B~JO fails the schema for the opposite reason: it emits all ten fields but echoes the input transaction as an eleventh key in 68.8\% of records, leaving 18.1\% schema-valid at 0.698 F1. R-EM can likewise be nonzero when omitted keys match empty reference fields. After matched fine-tuning, all 16 configurations parse on 95.8--100\% and satisfy the schema on 95.6--100\% of records. Adaptation therefore improves both extraction and output-contract compliance, while the FT--JO contrast remains a compound prompt comparison rather than an isolated reasoning effect.

\subsection{Scale and Adapter Rank}
\label{sec:scaling}

\begin{figure}[t]
\centering
\includegraphics[width=\columnwidth]{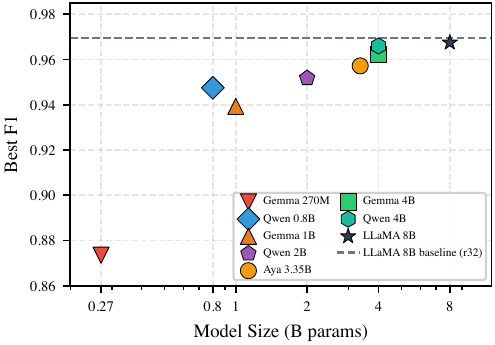}
\caption{Model size vs.\ best F1 score. Within Gemma~3, F1 rises with size and flattens above 1B; cross-family points are not size-controlled.}
\label{fig:scaling}
\end{figure}

Fig.~\ref{fig:scaling} shows F1 scaling with model size. Within Gemma~3 (same architecture at 270M, 1B, 4B), gains are +6.58 points from 270M$\to$1B and +2.31 from 1B$\to$4B, exhibiting diminishing returns above ${\sim}$1B.

The Qwen~3.5 0.8B model (F1=94.75\% with FT Nothink) nearly matches the 2B variant (95.18\%) and exceeds Gemma~3~1B (93.93\%), a model $25\%$ larger with a conventional transformer architecture. This observation is consistent with the hypothesis that Qwen~3.5's hybrid Mamba-attention design is effective for identifying field values in noisy input, but it does not isolate architecture: the families also differ in pretraining data, tokenizer, and instruction-tuning recipe. At the 4B scale, the best Qwen~3.5 variant (JO, F1=96.60\%) also surpasses Gemma~3~4B (FT, 96.24\%). Controlled pretraining ablations would be required to attribute either gap to the hybrid design.

\label{sec:lora_rank}

\textbf{Adapter rank.} We compare LoRA rank~8 against rank~32 on two models. On Aya~3.35B, rank~32 (60.5M trainable parameters, 1.8\%) achieves F1=95.29\% vs.\ rank~8's 95.22\% (both JO prompt), a small observed difference despite 4$\times$ more adapter parameters. This is not a fully controlled rank ablation: the Aya rank-32 run uses learning rate $5\times10^{-5}$, whereas rank~8 uses the standardized $10^{-4}$ setting, so the difference cannot be attributed to rank alone. On LLaMA~3.1-8B, our rank-8 FT fine-tune reaches 96.75\% F1 compared with the separately trained production reference's 96.95\%, a gap of 0.20 points. The retained reference artifact verifies the scores but not the historical adapter metadata, so its rank~32 attribution is author-attested. The two comparisons therefore support rank~8 as a practical default in our tested settings, not a causal claim that rank is irrelevant. For Aya, rank~8 reduces adapter storage by approximately 4$\times$; the absent production-reference adapter prevents independently verifying the corresponding LLaMA storage ratio.

\subsection{Training Dynamics}

\begin{figure}[t]
\centering
\includegraphics[width=0.95\columnwidth]{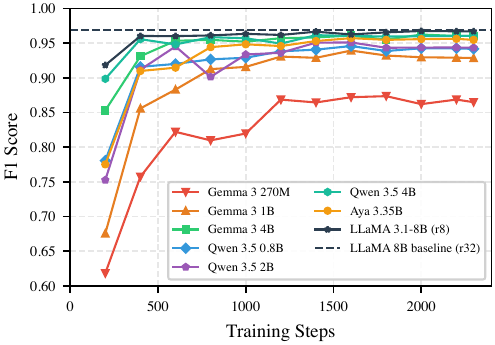}
\caption{F1 progression during training (FT prompt). Dashed: rank-32 LLaMA~8B production reference.}
\label{fig:training}
\end{figure}

Fig.~\ref{fig:training} shows F1 across training steps. Larger models reach high F1 earlier (Gemma~4B reaches 95\% by step~600 vs.\ 1,200+ for 270M), and our LLaMA~3.1-8B rank-8 fine-tune approaches the rank-32 production reference (dashed line) within the first 1,000 steps and tracks it closely thereafter. Oracle test maxima typically occur at steps 1,400--2,000 (epochs 3.7--5.2), with lower scores at some later checkpoints.

\subsection{Prompt and Training-Template Comparison}
\label{sec:prompt}

\begin{figure}[t]
\centering
\includegraphics[width=\columnwidth]{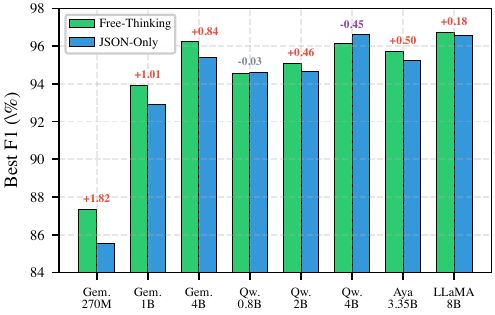}
\caption{Free-Thinking vs.\ JSON-Only. FT leads for six of eight models; Qwen~3.5~4B and, marginally, Qwen~3.5~0.8B favor JO.}
\label{fig:prompt}
\end{figure}

Fig.~\ref{fig:prompt} compares Free-Thinking and JSON-Only across model sizes. FT outperforms JO for most models, with gains from $+0.18$ points (LLaMA~8B) and $+0.46$ (Qwen~2B) up to $+1.82$ points (Gemma~270M); the gap is most pronounced for Gemma and Aya. The variants share output labels and entity rules, but FT jointly adds an internal-analysis instruction, an explicit ten-key schema, and 830 prompt characters. The experiment therefore establishes that this compound FT prompt helps most tested models, not that chain-of-thought alone causes the gain; the pattern is nonetheless directionally consistent with reports that reasoning-style instructions can help without reasoning appearing in outputs~\cite{deng2024explicit}. The larger observed gaps for smaller Gemma models are suggestive, but a factorial prompt ablation would be required to separate reasoning framing, schema explicitness, and length.

The Qwen~3.5~4B model reverses this trend---JO (F1=96.60\%) outperforms FT (96.15\%) by 0.45 points, making it the best sub-8B model overall. It may benefit from JO's shorter, less explicit formulation, but the current compound comparison cannot identify why; a factorial prompt ablation is required.

\label{sec:think}

\textbf{Training templates.} Qwen~3.5 natively supports extended thinking via \texttt{<think>...</think>} tags. We compare two LLaMA-Factory training templates that differ in output control-token formatting:

\begin{itemize}
    \item \textbf{Think}: Uses the reasoning template; every training label contains an empty \texttt{<think>...</think>} block before the JSON object.
    \item \textbf{Nothink}: Uses the base template---excludes thought tokens from training labels; inference prompt instructs direct JSON output.
\end{itemize}

Each template has its own training configuration (separate checkpoint directories), ensuring a fair comparison. We evaluate all six Qwen~3.5 configurations (3 sizes $\times$ 2 prompts) under both templates.

\textbf{Think-trained models reproduce the supervised empty block.} Across all Think-trained Qwen variants and the 1,213 common-evaluation records, every prediction contains an empty block (\texttt{<think>$\backslash$n$\backslash$n</think>}) followed immediately by JSON, matching the label format. Thus this experiment compares empty reasoning-control tokens with no thought tags; it does not test substantive reasoning supervision or show that models autonomously learned to suppress reasoning.

\textbf{Accuracy is near-identical across templates.} Table~\ref{tab:think_nothink} shows that Think and Nothink produce F1 scores within 0.004 of each other for every configuration. For FT prompts, Nothink matches or slightly exceeds Think at all three model sizes ($\Delta \leq$0.002). For JO prompts, results are mixed: Qwen~4B JO Think achieves the overall best F1 (0.9660 vs.\ Nothink's 0.9624), while Qwen~2B JO Nothink leads by +0.002.

\begin{table}[t]
\centering
\caption{Think vs.\ Nothink for Qwen~3.5 (best-checkpoint F1, 1,213 samples). Speed is the median of repeated batch-4 evaluations of the selected checkpoint on DGX Spark. $\Delta$F1 is Think minus Nothink.}
\label{tab:think_nothink}
\footnotesize
\begin{tabular}{@{}llccccr@{}}
\toprule
& & \multicolumn{2}{c}{\textbf{Think}} & \multicolumn{2}{c}{\textbf{Nothink}} & \\
\cmidrule(lr){3-4} \cmidrule(lr){5-6}
\textbf{Model} & \textbf{Pr.} & F1 & s/s & F1 & s/s & $\Delta$F1 \\
\midrule
Qwen 3.5 0.8B & FT & .9458 & 1.73 & .9475 & 1.86 & $-$.002 \\
Qwen 3.5 0.8B & JO & .9461 & 1.88 & .9452 & 2.01 & +.001 \\
Qwen 3.5 2B & FT & .9510 & 1.21 & .9518 & 1.27 & $-$.001 \\
Qwen 3.5 2B & JO & .9464 & 1.28 & .9486 & 1.34 & $-$.002 \\
Qwen 3.5 4B & FT & .9615 & 0.55 & .9624 & 0.59 & $-$.001 \\
Qwen 3.5 4B & JO & .9660 & 0.61 & .9624 & 0.62 & +.004 \\
\bottomrule
\end{tabular}
\end{table}

\textbf{Nothink is slightly faster.} Every Nothink variant runs 1.02--1.08$\times$ its Think counterpart (e.g., 0.8B FT: 1.86 vs.\ 1.73~s/s; 4B FT: 0.59 vs.\ 0.55~s/s). The gap is small but consistent in sign across all six pairs, as expected when the omitted thought block is short.


\subsection{Per-Field Performance}

\begin{figure*}[t]
\centering
\includegraphics[width=\textwidth]{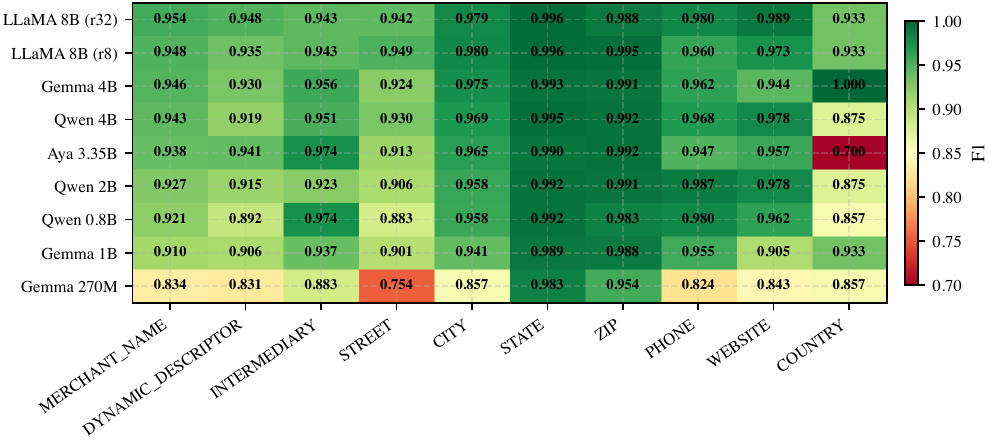}
\caption{Per-field F1 across models (FT prompt, best checkpoint). LLaMA rows are the rank-32 production reference (r32) and our rank-8 re-fine-tune (r8). Fields ordered by difficulty; models by overall F1.}
\label{fig:heatmap}
\end{figure*}

Fig.~\ref{fig:heatmap} shows per-field F1 across models, with the LLaMA~8B rank-32 production reference (r32) and our rank-8 study run (r8) broken out as separate rows. \textsc{State} (0.983--0.996) and \textsc{Zip} (0.954--0.995) score very high even at 270M. \textsc{Country} shows the widest spread (0.700--1.000), but rests on only eight evaluation records, and its tokens sometimes sit inside the merchant name or street. \textsc{Merchant\_Name} (0.834--0.954) and \textsc{Street} (0.754--0.949) degrade sharply below 1B. The two LLaMA~8B rows track each other closely across every field (within 0.02 in all cases, within 0.01 for seven of ten fields), reinforcing the rank-8-vs-rank-32 finding: the rank-8 study adapter loses very little per-field relative to the production reference. Appendix~\ref{app:perfield} tabulates per-field scores for each model's best-scoring variant.

\subsection{Accuracy--Throughput Pareto Analysis}
\label{sec:pareto}

\begin{figure}[t]
\centering
\includegraphics[width=\columnwidth]{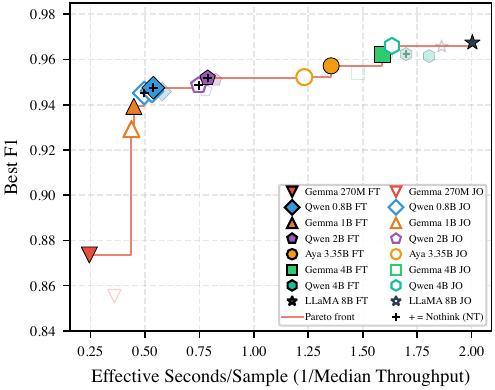}
\caption{Accuracy--throughput Pareto analysis. The horizontal axis is effective seconds per sample (inverse median batched throughput), not request latency. Pareto-optimal configurations define the frontier; dominated variants are faded.}
\label{fig:pareto}
\end{figure}

Fig.~\ref{fig:pareto} plots best F1 against effective seconds per sample, the inverse of median batched throughput. The frontier spans from Gemma~3~270M~FT (0.24\,s, F1=87.35\%) through Gemma~3~1B, the Qwen~3.5~0.8B and 2B Nothink variants, Qwen~3.5~0.8B~JO, Aya~3.35B, and Gemma~3~4B~FT to Qwen~3.5~4B~JO (1.63\,s, F1=96.60\%) and LLaMA~8B~FT (2.00\,s, F1=96.75\%). Nine of the twenty-two configurations are dominated, among them LLaMA~8B~JO, Gemma~3~4B~JO, and every Qwen~3.5~4B variant except JO. Under this matched measurement, switching from LLaMA~8B~FT (0.50 samples/s) to Qwen~3.5~4B~JO (0.61 samples/s) trades 0.15 F1 points for a $1.2\times$ lower inverse-throughput estimate. This ratio should not be read as a request-latency guarantee.

\section{Production Serving Validation on Databricks}
\label{sec:serving}

The results above characterize models under local, batched inference on a single DGX Spark workstation. Production deployment instead exposes each model through a managed serving layer---containerized endpoints, request routing, and per-request rather than batched execution---which can perturb observed accuracy. Because our study is deployment-oriented, we verify that its conclusions survive this transition: we register fine-tuned variants to the MLflow Model Registry, deploy each as a Databricks Model Serving endpoint, and re-evaluate the endpoint on the common 1,213-record evaluation subset. Endpoints are served via the standard HuggingFace transformers~\cite{wolf2020transformers} library. We deploy all 14 primary (family\,$\times$\,size\,$\times$\,prompt) fine-tuned variants spanning the three small-model families, including the study's top-ranked sub-8B model Qwen~3.5~4B~JO. Endpoint deployment of the LLaMA~3.1-8B variants (Table~\ref{tab:main_results}) is deferred to future work; the focus here is the sub-8B candidates that would actually replace the 8B in production.

Table~\ref{tab:databricks} compares each endpoint's F1 against the corresponding benchmark F1 from Table~\ref{tab:main_results}.

\begin{table}[t]
\centering
\caption{Databricks Model Serving endpoint accuracy for the 14 deployed variants. \textbf{Bench.\ F1} is the best-checkpoint score from Table~\ref{tab:main_results}; \textbf{Endpt.\ F1}~$=$~\textbf{Bench.\ F1}~$+~\Delta$F1. Rows ordered by $\Delta$F1.}
\label{tab:databricks}
\footnotesize
\begin{tabular}{@{}llrrr@{}}
\toprule
\textbf{Model} & \textbf{Pr.} & \textbf{Bench.\ F1} & \textbf{Endpt.\ F1} & \textbf{$\Delta$F1} \\
\midrule
Qwen 3.5 2B   & JO & .9464 & .9487 & $+$.0023 \\
Gemma 3 4B    & JO & .9540 & .9547 & $+$.0007 \\
Qwen 3.5 4B   & JO & .9660 & .9657 & $-$.0003 \\
Qwen 3.5 0.8B & JO & .9461 & .9453 & $-$.0008 \\
Qwen 3.5 0.8B & FT & .9458 & .9442 & $-$.0016 \\
Qwen 3.5 4B   & FT & .9615 & .9592 & $-$.0023 \\
Qwen 3.5 2B   & FT & .9510 & .9485 & $-$.0025 \\
Gemma 3 4B    & FT & .9624 & .9592 & $-$.0032 \\
Gemma 3 1B    & JO & .9292 & .9246 & $-$.0046 \\
Gemma 3 270M  & JO & .8553 & .8502 & $-$.0051 \\
Gemma 3 270M  & FT & .8735 & .8665 & $-$.0070 \\
Gemma 3 1B    & FT & .9393 & .9285 & $-$.0108 \\
Aya 3.35B     & JO & .9522 & .9250 & $-$.0272 \\
Aya 3.35B     & FT & .9572 & .9063 & $-$.0509 \\
\midrule
\multicolumn{4}{@{}l}{\textbf{Mean $\Delta$F1}} & \textbf{$-$.0081} \\
\bottomrule
\end{tabular}
\end{table}

\textbf{Benchmark accuracy transfers to production serving.} Across the 14 endpoints the mean change is only $-0.0081$~F1. Excluding the two Aya~3.35B variants (discussed below), every endpoint falls within 1.1 points of its benchmark score, and two---Qwen~3.5~2B~JO ($+0.0023$) and Gemma~3~4B~JO ($+0.0007$)---marginally \emph{exceed} it. This confirms the central premise of the study: the accuracy rankings established under local evaluation hold when the models are deployed as managed production endpoints.

\textbf{Aya~3.35B is the sole exception.} Both Aya variants degrade markedly once served---FT by 5.1 points ($.9572\rightarrow.9063$) and JO by 2.7 points ($.9522\rightarrow.9250$)---far larger than any Gemma or Qwen variant, whose worst case is a 1.1-point drop. The family-specific pattern is consistent with a serving-stack compatibility issue for the Cohere2-based model, but the endpoint study does not isolate the cause; runtime versions, serialization, and request handling remain potential confounds. Aya is therefore a weaker choice in this tested serving setup despite competitive local accuracy, reinforcing the need to validate each candidate on its target stack.

\section{Deployment Recommendations}
\label{sec:deployment}

Based on our results, we propose tiered deployment strategies:

\textbf{Tier 1 --- Maximum accuracy} ($>$96\% F1): for sites willing to keep the 8B footprint, the reproducible LLaMA~3.1-8B rank-8 FT fine-tune (96.75\% F1) is an adapter-efficient candidate; endpoint validation of this variant is deferred to future work. For a smaller alternative, Qwen~3.5~4B JO at 96.60\% F1 closes to within 0.35 points of the production reference at half the parameter count. Gemma~3~4B FT or Qwen~3.5~4B Nothink variants achieve 96.24\% and offer further deployment flexibility.

\textbf{Tier 2 --- Balanced} ($>$95\% F1): Qwen~3.5~2B FT Nothink (95.18\% F1, 1.27~s/s) is the recommended local configuration, combining accuracy and throughput. This Nothink variant was not among the 14 primary endpoint deployments; the corresponding Think FT endpoint changed by $-0.0025$ F1 in Table~\ref{tab:databricks}, so direct Nothink endpoint validation remains future work. Aya~3.35B (95.72\% F1, 14.5\,h training, 0.74~s/s) edges it locally but loses 3--5 F1 points in the tested Databricks setup; we therefore do not recommend Aya for that serving stack despite its local score.

\textbf{Tier 3 --- Throughput-critical} ($>$94\% F1): Qwen~3.5~0.8B FT Nothink achieves 94.75\% F1 at 1.86 s/s on DGX Spark. Its observed accuracy--throughput trade-off makes it the preferred tested option for edge or high-throughput deployments.

\textbf{Prompt strategy.} In the tested compound variants, Free-Thinking provides +0.5--1.8 F1 points for Gemma and Aya, while JSON-Only yields the best Qwen~3.5~4B result (96.60\%); smaller Qwen models are comparable. Because FT also adds an explicit schema and length, we recommend selecting between these complete prompts empirically rather than attributing the difference to reasoning alone.

\textbf{Training template.} For Qwen~3.5, Nothink is a practical default because it matches Think accuracy closely and simplifies parsing by omitting empty thought tags. Its FT runs also show higher throughput here, although timing variability prevents a causal attribution to the tags.

\textbf{LoRA configuration.} Rank~8 is a practical default in our experiments: Aya rank~32 yields only +0.07 points at approximately 4$\times$ the adapter size, while the LLaMA rank-8 study run is within 0.20 points of the production reference attributed by internal records to rank~32. 

\textbf{Production serving.} Endpoint validation (Section~\ref{sec:serving}) finds a mean $\Delta$F1 of $-0.0029$ across the Gemma and Qwen endpoints; the all-endpoint mean including Aya is $-0.0081$. Aya~3.35B is the exception, losing 3--5 F1 points in this setup. We therefore favor the tested Gemma~3 and Qwen~3.5 endpoints and advise validating any candidate directly on its target stack.

\textbf{Experiment-then-deploy workflow.} Our results support a deliberate two-stage workflow: \emph{experiment and select} with LLaMA-Factory on flexible compute (DGX Spark in our case), then \emph{deploy and serve} only the selected adapters via MLflow on a security-constrained production cluster. This separation lets teams evaluate any newly released open-source LLM as soon as its weights become available, while keeping the production attack surface limited to a small set of vetted models. It also keeps experimentation cost predictable, and we expect the pattern to generalize to other enterprise settings with similar tooling and policy constraints.

\section{Limitations and Generalizability}
\label{sec:limitations}

This study evaluates one financial extraction schema and one labeled corpus; transfer to KYC parsing, remittance text, invoices, addresses, or other domains has not been established. The rank, compound-prompt, and compact-model findings are therefore hypotheses for other schema-constrained extraction tasks, whereas absolute F1, R-EM, throughput, and checkpoint locations are setting-specific. Model-family comparisons confound architecture with pretraining data, tokenizer, and instruction tuning; FT versus JO confounds reasoning framing, explicit schema, and prompt length; and Think versus Nothink compares empty control-token blocks rather than substantive reasoning traces. We include no CRF or encoder-based baseline and claim no superiority over traditional NER systems: a faithful sequence-labeling comparison needs the JSON-to-BIO conversion and post-processing noted in Section~\ref{sec:related}, which we leave to future work after the split is publicly released. Section~\ref{sec:zeroshot} instead evaluates adaptation through the complete 16-configuration zero-shot sweep.
Serving validation covers one managed stack, excludes the 8B variants and Nothink endpoints, and does not causally identify Aya's degradation. Finally, local throughput is measured on a single DGX Spark at batch size~4 under one software stack; inverse throughput is not request latency. Repeating the study across tasks, random seeds, hardware, and serving runtimes is necessary before treating the observed trade-offs as general laws.

\section{Conclusion}
\label{sec:conclusion}

We presented a deployment-oriented study of LoRA-fine-tuned small language models for financial entity extraction, evaluating 23 retained fine-tuning runs from 0.27B to 8B parameters plus a separately trained production reference across two compound prompts and two training templates. The rank-8 LLaMA~3.1-8B fine-tune achieves 96.75\% F1, within 0.20 points of the rank-32 production reference. The best sub-8B model, Qwen~3.5~4B with JSON-Only prompting, achieves 96.60\% F1 and 91.67\% strict R-EM with a $1.2\times$ lower inverse-throughput estimate than rank-8 LLaMA~3.1-8B~FT. Qwen~3.5~0.8B reaches 94.75\% F1, although the cross-family design does not isolate the source of that result. Think and Nothink templates produce near-identical Qwen accuracy; Nothink omits the supervised empty thought tags and simplifies parsing. Across 14 Databricks endpoints, local scores generally transfer, while Aya degrades materially for reasons not isolated here. Together, the results support an experiment-then-deploy workflow and show that compact adapters and models are credible candidates for this workload, subject to validation-based checkpoint selection and direct testing on the target serving stack.

\section*{Reproducibility}
Upon internal approval, a public snapshot of the training and evaluation package will be released on GitHub\footnote{\url{https://github.com/inflaton/gaime-slm}}. The snapshot will include the 6,125/668/1,222 dataset splits and 1,213-record common evaluation index; the Free-Thinking and JSON-Only prompts; LLaMA-Factory configurations for the 23 retained fine-tuning runs; local prediction and scoring code for canonicalized field F1 and strict R-EM; all 12 stored checkpoint metrics for 22 runs and the retained checkpoint for Aya rank~32; and the per-request managed-endpoint exports for the 14 deployed variants, from which the endpoint scores in Table~\ref{tab:databricks} can be reproduced with the same evaluator on the same 1,213-record subset.
The snapshot's README will document the environment and commands needed to reproduce the local results.

\appendices

\section{Prompt Templates}
\label{app:prompts}

We provide both prompt templates used for fine-tuning, abridged: three descriptor examples are elided from the shared rules. The two prompts share the same entity classification rules but differ in two places: (i)~the preamble instruction, and (ii)~the output specification---the FT prompt includes an explicit JSON schema, whereas the JO prompt omits it and relies on the entity keys enumerated in the rules to elicit correct JSON. The transaction string is appended after the \texttt{TRANSACTION:} token.

\subsection{Free-Thinking (FT) Prompt}

\begin{small}
\begin{verbatim}
### Instructions
Please follow the two-step approach below.
#### Step 1 (Free-Form Analysis)
1. Analyze the bank transaction on your own
   without applying any rules or guidelines.
2. Provide a free-form interpretation of the
   transaction, focusing on any entities you
   naturally identify.
3. Perform this internally, do not return this
   as an output.
#### Step 2 (Rule-Based Classification)
1. Re-analyze the same transaction strictly
   using the Entity Identification and
   Classification Rules below.
2. Return a JSON object containing exactly the
   keys listed under JSON Output Format.
3. Leave any key as an empty string if you do
   not find a corresponding value.
4. Include no explanations or additional text
   beyond the JSON object.
\end{verbatim}
\end{small}

\noindent Followed by the shared entity classification rules (Appendix~\ref{app:rules}) and the JSON output format with explicit schema (Appendix~\ref{app:output}).

\subsection{JSON-Only (JO) Prompt}

\begin{small}
\begin{verbatim}
Extract entities from the following bank
transaction string and return a JSON object.
\end{verbatim}
\end{small}

\noindent Followed by the same shared entity classification rules (Appendix~\ref{app:rules}) and an output instruction for the JSON schema (Appendix~\ref{app:output}).

\subsection{Shared Entity Classification Rules}
\label{app:rules}

Both prompts include identical classification rules:

\begin{small}
\begin{verbatim}
1. General Approach:
 - Extract entities without assumptions
   or constraints.
 - Identify all possible segments and
   interpret them contextually.
 - Consider multiple interpretations for
   ambiguous segments.
2. Entity Classification Rules:
 - INTERMEDIARY: Payment facilitators or
   intermediaries (e.g., PayPal, Square,
   Toast, Klarna, DoorDash, CashApp).
   Often appears before MERCHANT_NAME.
   Example: "SQ* WAL-MART #1234"
     -> {"INTERMEDIARY": "SQ*",
         "MERCHANT_NAME": "WAL-MART #1234"}
 - MERCHANT_NAME: Primary business where
   the transaction occurred. May include
   store numbers (e.g., "WAL-MART#6732",
   "STARBUCKS 27057", "TARGET 0001332904",
   "SHELL OIL 12679923008"). If a brand
   name is immediately followed by
   product/service text, include the
   entire portion (e.g., "MICROSOFT
   * ONE DRIVE").
 - DYNAMIC_DESCRIPTOR: Transaction-
   specific identifiers (order numbers,
   reference codes, invoice details).
   Examples: "*INV-5384", "-X99S90S43",
     "* O #03901", "KG9RCW",
     "* HM3W5MY95J", "*RV8KK6372"
 - STREET: Street address; starts with a
   number or building identifier plus
   street name and known suffixes
   (e.g., ST, AVE, BLVD). Example:
   "2323 West 5th AvenueSte 230"
 - ZIP: 5- or 9-digit postal codes.
 - CITY: Recognizable US city or town
   names.
 - STATE: Two-letter US state code
   (e.g., "VA").
 - COUNTRY: Two- or three-letter code
   (e.g., "USA", "US").
 - PHONE: Phone numbers in any format
   (e.g., "800-365-6868", "4029357733",
   "+14086348696", "(800)262-3246").
 - WEBSITE: URL associated with the
   merchant (e.g., "AMZN.COM/BILL",
   "G.CO/HELPPAY#",
   "HTTPSWWW.DOORDASH.COM").
3. Handling Special Cases:
 - If a token might fit multiple
   categories, use position/formatting
   to determine the correct entity type.
 - If unsure, use your best guess.
 - If duplicate or partially duplicate
   information (e.g., website, street,
   city, state, country) appears at the
   end of MERCHANT_NAME, remove it and
   assign it to the appropriate key.
 - Omit complete and partial duplicate
   strings.
\end{verbatim}
\end{small}

\subsection{JSON Output}
\label{app:output}

The two prompts diverge in how they describe the output. The FT prompt includes an explicit JSON schema:

\begin{small}
\begin{verbatim}
### JSON Output Format
Return your Step 2 classification with
exactly these keys and nothing else:
{
  "INTERMEDIARY": "",
  "MERCHANT_NAME": "",
  "DYNAMIC_DESCRIPTOR": "",
  "STREET": "",
  "ZIP": "",
  "CITY": "",
  "STATE": "",
  "COUNTRY": "",
  "PHONE": "",
  "WEBSITE": ""
}
\end{verbatim}
\end{small}

\noindent The JO prompt omits the schema and ends with an instruction only:

\begin{small}
\begin{verbatim}
### Output Format
Return a JSON object with exactly these
keys. Leave any key as an empty string
if no value is found. Return only the
JSON object with no additional text.
\end{verbatim}
\end{small}

\noindent Models fine-tuned on the JO prompt still produce well-formed JSON: the ten entity keys are enumerated as bullet items in the Entity Classification Rules above (Appendix~\ref{app:rules}), so the model has the full key set even without an explicit schema.

\section{Training Configuration}
\label{app:training}

All models are fine-tuned using LLaMA-Factory~\cite{zheng2024llamafactory}. Below is the representative YAML configuration (shown for Qwen~3.5~4B with default qwen3\_5 template):
\begin{small}
\begin{verbatim}
model_name_or_path: Qwen/Qwen3.5-4B
stage: sft
do_train: true
finetuning_type: lora
lora_rank: 8
lora_alpha: 16
learning_rate: 1.0e-4
num_train_epochs: 6
per_device_train_batch_size: 4
gradient_accumulation_steps: 4
lr_scheduler_type: cosine
warmup_ratio: 0.1
save_steps: 200
do_sample: false
temperature: 0.0
template: qwen3_5  # or qwen3_5_nothink

\end{verbatim}
\end{small}


\section{Per-Field Detailed Results}
\label{app:perfield}

Table~\ref{tab:perfield_detail} provides per-field F1 scores for each model at its best-scoring variant. This complements the heatmap in Fig.~\ref{fig:heatmap}, which shows FT Think results only.

\begin{table}[!ht]
\centering
\caption{Per-field F1 at best checkpoint, one row per model: its best-scoring variant. LLaMA rows are the rank-32 production reference (r32) and our rank-8 re-fine-tune (r8). \textsc{Micro} is the support-weighted (micro-averaged) F1 from Table~\ref{tab:main_results}; rows are sorted by it. NT = Nothink.}
\label{tab:perfield_detail}
\scriptsize
\setlength{\tabcolsep}{2.1pt}
\begin{tabular}{@{}llccccccccccc@{}}
\toprule
\textbf{Model} & \textbf{Pr.} & \rotatebox{90}{\textsc{Micro}} & \rotatebox{90}{\textsc{Merch.}} & \rotatebox{90}{\textsc{Dyn.D.}} & \rotatebox{90}{\textsc{Inter.}} & \rotatebox{90}{\textsc{Street}} & \rotatebox{90}{\textsc{City}} & \rotatebox{90}{\textsc{State}} & \rotatebox{90}{\textsc{Zip}} & \rotatebox{90}{\textsc{Phone}} & \rotatebox{90}{\textsc{Web.}} & \rotatebox{90}{\textsc{Ctry.}} \\
\midrule
LLaMA 8B r32 & FT & .970 & .954 & .948 & .943 & .942 & .979 & .996 & .988 & .980 & .989 & .933 \\
LLaMA 8B r8 & FT & .968 & .948 & .935 & .943 & .949 & .980 & .996 & .995 & .960 & .973 & .933 \\
Qwen 4B & JO & .966 & .949 & .938 & .956 & .935 & .970 & .996 & .995 & 1.00 & .967 & .933 \\
Gemma 4B & FT & .962 & .946 & .930 & .956 & .924 & .975 & .993 & .991 & .962 & .944 & 1.00 \\
Aya 3.35B & FT & .957 & .938 & .941 & .974 & .913 & .965 & .990 & .992 & .947 & .957 & .700 \\
Qwen 2B & FT NT & .952 & .929 & .899 & .933 & .907 & .959 & .994 & .988 & .973 & .995 & 1.00 \\
Qwen 0.8B & FT NT & .948 & .923 & .902 & .969 & .899 & .955 & .990 & .984 & .986 & .956 & .857 \\
Gemma 1B & FT & .939 & .910 & .906 & .937 & .901 & .941 & .989 & .988 & .955 & .905 & .933 \\
Gemma 270M & FT & .874 & .834 & .831 & .883 & .754 & .857 & .983 & .954 & .824 & .843 & .857 \\
\bottomrule
\end{tabular}
\end{table}

\section*{Acknowledgment}
This research was supported by the Singapore Ministry of Education (MOE) Academic Research Fund (AcRF) Tier 1 Grant (Grant No.~24-SIS-SMU-131).

\balance

\bibliographystyle{IEEEtran}
\bibliography{references}

@inproceedings{hu2022lora,
  title={{LoRA}: Low-Rank Adaptation of Large Language Models},
  author={Hu, Edward J and Shen, Yelong and Wallis, Phillip and Allen-Zhu, Zeyuan and Li, Yuanzhi and Wang, Shean and Wang, Lu and Chen, Weizhu},
  booktitle={International Conference on Learning Representations},
  year={2022}
}

@article{dettmers2023qlora,
  title={{QLoRA}: Efficient Finetuning of Quantized Language Models},
  author={Dettmers, Tim and Pagnoni, Artidoro and Holtzman, Ari and Zettlemoyer, Luke},
  journal={Advances in Neural Information Processing Systems},
  volume={36},
  year={2023}
}

@inproceedings{zhang2023adalora,
  title={{AdaLoRA}: Adaptive Budget Allocation for Parameter-Efficient Fine-Tuning},
  author={Zhang, Qingru and Chen, Minshuo and Bukharin, Alexander and He, Pengcheng and Cheng, Yu and Chen, Weizhu and Zhao, Tuo},
  booktitle={International Conference on Learning Representations},
  year={2023}
}

@article{team2024gemma,
  title={{Gemma} 3 Technical Report},
  author={{Gemma Team} and Kamath, Aishwarya and Ferret, Johan and Pathak, Shreya and Vieillard, Nino and Merhej, Ramona and Perrin, Sarah and Matejovicova, Tatiana and Ram{\'e}, Alexandre and Rivi{\`e}re, Morgane and others},
  journal={arXiv preprint arXiv:2503.19786},
  year={2025}
}

@misc{yang2024qwen2,
  title={{Qwen3.5}: Towards Native Multimodal Agents},
  author={{Qwen Team}},
  month={February},
  year={2026},
  howpublished={Qwen blog},
  url={https://qwen.ai/blog?id=qwen3.5}
}

@article{ustun2024aya,
  title={Tiny {Aya}: Bridging Scale and Multilingual Depth},
  author={Salamanca, Alejandro R. and Abagyan, Diana and D'souza, Daniel and Khairi, Ammar and Mora, David and Dash, Saurabh and Aryabumi, Viraat and Rajaee, Sara and Mofakhami, Mehrnaz and Sahu, Ananya and others},
  journal={arXiv preprint arXiv:2603.11510},
  year={2026}
}

@article{dubey2024llama,
  title={The {Llama} 3 Herd of Models},
  author={Dubey, Abhimanyu and Jauhri, Abhinav and Pandey, Abhinav and others},
  journal={arXiv preprint arXiv:2407.21783},
  year={2024}
}

@article{gu2023mamba,
  title={Mamba: Linear-Time Sequence Modeling with Selective State Spaces},
  author={Gu, Albert and Dao, Tri},
  journal={arXiv preprint arXiv:2312.00752},
  year={2023}
}

@article{dao2024transformers,
  title={Transformers are {SSMs}: Generalized Models and Efficient Algorithms Through Structured State Space Duality},
  author={Dao, Tri and Gu, Albert},
  journal={arXiv preprint arXiv:2405.21060},
  year={2024}
}

@inproceedings{lample2016neural,
  title={Neural Architectures for Named Entity Recognition},
  author={Lample, Guillaume and Ballesteros, Miguel and Subramanian, Sandeep and Kawakami, Kazuya and Dyer, Chris},
  booktitle={Proc.\ NAACL-HLT},
  pages={260--270},
  year={2016},
  doi={10.18653/v1/N16-1030}
}

@inproceedings{ma2016end,
  title={End-to-end Sequence Labeling via Bi-directional {LSTM-CNNs-CRF}},
  author={Ma, Xuezhe and Hovy, Eduard},
  booktitle={Proc.\ 54th Annu.\ Meeting Assoc.\ Comput.\ Linguistics},
  pages={1064--1074},
  year={2016},
  doi={10.18653/v1/P16-1101}
}

@article{wang2023gptner,
  title={{GPT-NER}: Named Entity Recognition via Large Language Models},
  author={Wang, Shuhe and Sun, Xiaofei and Li, Xiaoya and Ouyang, Rongbin and Wu, Fei and Zhang, Tianwei and Li, Jiwei and Wang, Guoyin},
  journal={arXiv preprint arXiv:2304.10428},
  year={2023}
}

@article{li2023evaluating,
  title={Evaluating {ChatGPT}'s Information Extraction Capabilities: An Assessment of Performance, Explainability, Calibration, and Faithfulness},
  author={Li, Bo and Fang, Gexiang and Yang, Yang and Wang, Quansen and Ye, Wei and Zhao, Wen and Zhang, Shikun},
  journal={arXiv preprint arXiv:2304.11633},
  year={2023}
}

@article{wei2022chain,
  title={Chain-of-Thought Prompting Elicits Reasoning in Large Language Models},
  author={Wei, Jason and Wang, Xuezhi and Schuurmans, Dale and Bosma, Maarten and Ichter, Brian and Xia, Fei and Chi, Ed and Le, Quoc V and Zhou, Denny},
  journal={Advances in Neural Information Processing Systems},
  volume={35},
  pages={24824--24837},
  year={2022}
}

@article{du2025nlpfinance,
  title={Natural language processing in finance: A survey},
  author={Du, Kelvin and Zhao, Yazhi and Mao, Rui and others},
  journal={Information Fusion},
  volume={115},
  pages={102755},
  year={2025},
  publisher={Elsevier},
  doi={10.1016/j.inffus.2024.102755}
}

@article{shah2022flue,
  title={When {FLUE} Meets {FLANG}: Benchmarks and Large Pretrained Language Model for Financial Domain},
  author={Shah, Raj Sanjay and Chawla, Kunal and Eidnani, Dheeraj and others},
  journal={arXiv preprint arXiv:2211.00083},
  year={2022}
}

@inproceedings{strubell2019energy,
  title={Energy and Policy Considerations for Deep Learning in {NLP}},
  author={Strubell, Emma and Ganesh, Ananya and McCallum, Andrew},
  booktitle={Proc.\ 57th Annu.\ Meeting Assoc.\ Comput.\ Linguistics},
  pages={3645--3650},
  year={2019},
  doi={10.18653/v1/P19-1355}
}

@article{abdin2024phi3,
  title={Phi-3 Technical Report: A Highly Capable Language Model Locally on Your Phone},
  author={Abdin, Marah and Jacobs, Sam Ade and Awan, Ammar Ahmad and others},
  journal={arXiv preprint arXiv:2404.14219},
  year={2024}
}

@inproceedings{zheng2024llamafactory,
  title={{LlamaFactory}: Unified Efficient Fine-Tuning of 100+ Language Models},
  author={Zheng, Yaowei and Zhang, Richong and Zhang, Junhao and Ye, Yanhan and Luo, Zheyan and Ma, Zhangchi and Ma, Yongqiang},
  booktitle={Proc.\ ACL: Syst.\ Demonstrations},
  year={2024}
}

@article{deng2024explicit,
  title={From Explicit {CoT} to Implicit {CoT}: Learning to Internalize {CoT} Step by Step},
  author={Deng, Yuntian and Choi, Yejin and Shieber, Stuart},
  journal={arXiv preprint arXiv:2405.14838},
  year={2024}
}

@article{zhang2023multimodal,
  title={Multimodal Chain-of-Thought Reasoning in Language Models},
  author={Zhang, Zhuosheng and Zhang, Aston and Li, Mu and Zhao, Hai and Karypis, George and Smola, Alex},
  journal={arXiv preprint arXiv:2302.00923},
  year={2023}
}

@inproceedings{wolf2020transformers,
  title={{HuggingFace}'s {Transformers}: State-of-the-Art Natural Language Processing},
  author={Wolf, Thomas and Debut, Lysandre and Sanh, Victor and others},
  booktitle={Proc.\ EMNLP: Syst.\ Demonstrations},
  pages={38--45},
  year={2020},
  doi={10.18653/v1/2020.emnlp-demos.6}
}

@inproceedings{huang2025chinese,
  title={Optimizing {Chinese}-to-{English} Translation Using Large Language Models},
  author={Huang, Donghao and Wang, Zhaoxia},
  booktitle={Proc.\ 2025 IEEE Symposium on Computational Intelligence in Natural Language Processing and Social Media (CI-NLPSoMe Companion)},
  year={2025},
  address={Trondheim, Norway},
  publisher={IEEE},
  doi={10.1109/SSCI64329.2025.10970768}
}

@inproceedings{huang2025turtle,
  title={Logical Reasoning with {LLM}s via Few-Shot Prompting and Fine-Tuning: A Case Study on Turtle Soup Puzzles},
  author={Huang, Donghao and Wang, Zhaoxia},
  booktitle={Proc.\ 2025 IEEE Symposium on Computational Intelligence in Natural Language Processing and Social Media (CI-NLPSoMe Companion)},
  year={2025},
  address={Trondheim, Norway},
  publisher={IEEE},
  doi={10.1109/CI-NLPSoMeCompanion65206.2025.10977921}
}

@article{huang2026payments,
  title={Beyond Task Success: Measuring Workflow Fidelity in {LLM}-Based Agentic Payment Systems},
  author={Huang, Donghao and Chua, Joon Kiat and Wang, Zhaoxia},
  journal={arXiv preprint arXiv:2605.06457},
  year={2026}
}

@inproceedings{lafferty2001conditional,
  title={Conditional Random Fields: Probabilistic Models for Segmenting and Labeling Sequence Data},
  author={Lafferty, John and McCallum, Andrew and Pereira, Fernando C. N.},
  booktitle={Proc.\ 18th Int.\ Conf.\ Mach.\ Learn.},
  pages={282--289},
  year={2001}
}

@inproceedings{devlin2019bert,
  title={{BERT}: Pre-training of Deep Bidirectional Transformers for Language Understanding},
  author={Devlin, Jacob and Chang, Ming-Wei and Lee, Kenton and Toutanova, Kristina},
  booktitle={Proc.\ NAACL-HLT},
  pages={4171--4186},
  year={2019},
  doi={10.18653/v1/N19-1423}
}

\end{document}